# Data-Oriented Language Processing
# An Overview

Rens Bod and Remko Scha

Department of Computational Linguistics

Institute for Logic, Language and Computation

University of Amsterdam

Spuistraat 134, 1012 VB Amsterdam

The Netherlands

Rens.Bod@let.uva.nl

Remko.Scha@let.uva.nl

cmp-lg/9611003   14 Nov 1996

## Abstract

Data-oriented models of language processing embody the assumption that human language perception and production works with representations of concrete past language experiences, rather than with abstract grammar rules. Such models therefore maintain large corpora of linguistic representations of previously occurring utterances. When processing a new input utterance, analyses of this utterance are constructed by combining fragments from the corpus; the occurrence-frequencies of the fragments are used to estimate which analysis is the most probable one. This paper motivates the idea of data-oriented language processing by considering the problem of syntactic disambiguation. One relatively simple parsing/disambiguation model that implements this idea is described in some detail. This model assumes a corpus of utterances annotated with labelled phrase-structure trees, and parses new input by combining subtrees from the corpus; it selects the most probable parse of an input utterance by considering the sum of the probabilities of all its derivations. The paper discusses some experiments carried out with this model. Finally, it reviews some other models that instantiate the data-oriented processing approach. Many of these models also employ labelled phrase-structure trees, but use different criteria for extracting subtrees from the corpus or employ different disambiguation strategies; other models use richer formalisms for their corpus annotations.

## 1.  Introduction

During the last few years, a new approach to language processing has started to emerge, which has become known under various labels such as "data-oriented parsing", "corpus-based interpretation", and "tree-bank grammar" (cf. van den Berg et al. 1994; Bod 1992-96; Bod et al., 1996a/b; Bonnema 1996; Charniak 1996a/b; Goodman 1996; Kaplan 1996; Rajman 1995a/b; Scha 1990-92; Sekine & Grishman 1995; Sima'an et al., 1994; Sima'an 1995-96; Tugwell 1995). This approach, which we will call *data-oriented processing* or *DOP*, embodies the assumption that human language perception and production works with representations of concrete past language experiences, rather than with abstract linguistic rules. The models that instantiate this approach therefore maintain large corpora of linguistic representations of previously occurring utterances. When processing a new input utterance, analyses of this utterance are constructed by combining fragments from the corpus; the occurrence-frequencies of the fragments are used to estimate which analysis is the most probable one.

This paper motivates the idea of data-oriented language processing by considering the problem of syntactic disambiguation (section 1). One relatively simple parsing/disambiguation model that implements the approach is described in some detail, both mathematically (section 2) and computationally (section 3). This model assumes a corpus of utterances annotated with labelled phrase-structure trees, and extracts all (arbitrarily large and arbitrarily complex) subtrees from this corpus; it parses new input by combining these subtrees; it selects the most probable parse of an input utterance by considering the sum of the probabilities of all its derivations. The paper discusses some experiments carried out with this model (section 4). Finally, it reviews some other models that instantiate the data-oriented processing approach. Many of these models also employ labelled phrase-structure trees, but use different criteria for extracting subtrees from the corpus or employ different disambiguation strategies (section 5); other models use richer formalisms for their corpus annotations (section 6).

### 1.1  Competence grammars and performance models

Modern linguistic theory models natural languages after the formal languages of logic and mathematics: as well-defined infinite sets of sentence/meaning pairs, defined by consistent and non-redundant systems of formal rules ("competence grammars"). It conjectures that human minds actually employ such rule systems in producing and comprehending new utterances; at the same time, it acknowledges explicitly that a competence grammar alone cannot account for all aspects of human language processing: a person's language "performance" is also influenced by several other mental properties, that do not belong to the core-business of linguistics.

For instance, it is well-known that a person listening to a spoken language utterance may perceive one particular sentence, though a large set of grammatical word sequences are phonetically compatible with the incoming signal; and that a person may assign one particular meaning to an input sentence, though the competence grammar would allow many other possibilities. Such capabilities clearly require non-linguistic knowledge, concerning, for instance, the varying degrees of plausibility of different real-world situations, and the varying occurrence-likelihoods of different words and constructions and their meanings. Psycholinguistics and language technology must account for such capabilities, and thus face the challenge of actually embedding linguistic competence grammars in a theory of language performance.

So far, most psycholinguistic and language-technological work has indeed accepted the competence/performance division suggested by linguistic theory. The processes of language generation and comprehension have been assumed to make explicit use of the rules of the competence grammar. To explain the actual properties of language performance, one invokes the computational properties of the processes that access the competence rules, and their interactions with other knowledge sources.

This paper deals with a language-technological approach that breaks with this tradition. We conjecture that a person's knowledge of a language should not be modelled as a compact, non-redundant set of formal rules, but rather as a database that records the utterances the person experienced, represented as strings of words together with their syntactic structures, their (linguistic and extra-linguistic) contexts, and their meanings. Using such a database, a system may parse and interpret new input by constructing analogies with previously experienced sentence-analyses; no grammar rules that summarize linguistic regularities need to be invoked.

We will motivate and demonstrate this idea by considering one particular performance issue: the problem of syntactic disambiguation.

## 1.2  Syntactic disambiguation

As soon as a linguistic competence grammar is large enough to cover a non-trivial fragment of English, it assigns to many input sentences an extremely large number of alternative syntactic analyses. Human language users, however, tend to perceive only one or two of these. The combinatorial explosion of syntactic analyses (and corresponding semantic interpretations) of natural language sentences has been ignored by linguistic theory, but is well-recognized in psycholinguistics and language technology (cf. Church & Patil, 1983; MacDonald et al., 1994). Martin et al. (1983) list the number of different analyses their grammar assigns to some example sentences:

| | |
|---|---|
| List the sales of products in 1973. | 3 |
| List the sales of products produced in 1973. | 10 |
| List the sales of products in 1973 with the products in 1972. | 28 |
| List the sales of products produced in 1973 with the products produced in 1972. | 455 |

Because of the different attachment possibilities of prepositional phrases and relative clauses, a formal grammar must acknowledge many possible structures for such sentences. Human speakers of English, however, will fail to notice this dazzling degree of ambiguity; not more than a few analyses will spontaneously come to mind.

There are many different criteria that play a role in human disambiguation behavior. First of all we should note that syntactic disambiguation is to some extent a side-effect of *semantic* disambiguation. People prefer plausible *interpretations* to implausible ones -- where the plausibility of an interpretation is assessed with respect to the specific semantic/pragmatic context at hand, taking into account conventional world knowledge (which determines what beliefs and desires we tend to attribute to others), social conventions (which determine what beliefs and desires tend to get verbally expressed), and linguistic conventions (which determine *how* they tend to get verbalized).

For the remainder of this paper we will, however, mostly ignore the semantic dimension. We will focus instead on a very important non-semantic factor that influences human disambiguation behavior: the frequency of occurrence of lexical items and syntactic structures. It has been established that (1) people register frequencies and frequency-differences (e.g. Hasher & Chromiak, 1977; Kausler & Puckett, 1980; Pearlmutter & MacDonald, 1992), (2) analyses that a person has experienced before are preferred to analyses that must be newly constructed (e.g. Hasher & Zacks, 1984; Jacoby & Brooks, 1984; Fenk-Oczlon, 1989), and (3) this preference is influenced by the frequency of occurrence of analyses: more frequent analyses are preferred to less frequent ones (e.g. Fenk-Oczlon, 1989; Mitchell et al., 1992; Juliano & Tanenhaus, 1993; Gibson & Loomis, 1994).

These findings are not surprising -- they are predicted by general information-theoretical considerations. A system confronted with an ambiguous signal may optimize its behavior by taking into account which interpretations are more likely to be correct than others -- and past occurrence frequencies may be the most reliable indicator for these likelihoods.

### 1.3 Stochastic grammars

It is plausible therefore, that the human language processing system estimates the most probable analysis of a new input sentence, on the basis of successful analyses of previously encountered ones. But how is this done? What probabilistic information does the system derive from its past language experiences? The set of sentences that a language allows may best be viewed as infinitely large, and probabilistic information is used to compare alternative analyses of sentences never encountered before. A finite set of probabilities of units and combination operations must therefore be used to characterize an infinite set of probabilities of sentence-analyses.

This problem can only be solved if a more basic, non-probabilistic one is solved first: we need a characterization of the complete set of possible sentence-analyses of the language. That is exactly what the competence-grammars of theoretical syntax try to provide. Many probabilistic disambiguation models therefore build on that work: they characterize the probabilities of sentence-analyses by means of a "stochastic grammar", constructed out of a competence-grammar by augmenting the rules with application probabilities derived from a corpus. Different syntactic frameworks have been extended in this way. Examples are stochastic context-free grammar (Suppes, 1970; Sampson, 1986; Black et al., 1993), stochastic tree-adjoining grammar (Resnik, 1992; Schabes, 1992), and stochastic unification-based grammar (Briscoe & Carroll, 1993; Briscoe, 1994).

A statistically enhanced competence grammar of this sort defines all sentences of a language and all analyses of these sentences. It also assigns probabilities to each of these sentences and each of these analyses. It therefore makes definite predictions about an important class of performance phenomena: the preferences that people display when they must choose between different sentences (in language production and speech recognition), or between alternative analyses of sentences (in disambiguation).

The accuracy of these predictions, however, is necessarily limited. Stochastic grammars assume that the statistically significant language units coincide exactly with the lexical items and syntactic rules employed by the competence grammar. The most obvious case of frequency-based bias in human disambiguation behavior therefore falls outside their scope: the tendency to assign "normal" rather than innovative interpretations to platitudes and conventional phrases. Platitudes and conventional phrases demonstrate that syntactic constructions of arbitrary size and complexity may be statistically important, while they may be completely redundant from a purely linguistic point of view. The direct connection between competence grammar and stochastic disambiguation is therefore counterproductive, and should be abolished.

## 1.4  Data-Oriented Language Processing

The language experience of an adult language user consists of a large number of utterances. Each of these utterances contains a multitude of constructions: not only the tree of the whole sentence, and all its constituent trees, but also all patterns that can be extracted from these by introducing "free variables" for lexical elements or complex constituents. Which of these constructions are used in processing new input? Most stochastic grammars have assumed that the smallest constructions (the building blocks of a competence grammar) are the only relevant ones. We know that this is false, but it is not immediately clear what we should assume instead. We therefore adopt a framework which does not prejudge this issue, and which allows us to experiment with different assumptions.

It is obvious that a person's past language experience somehow determines the outcome of this person's sentence analysis processes. The basic idea behind the data-oriented approach is that this happens in an essentially unmediated way. As a representation of a person's past language experience, we use a large corpus of utterances with their analyses. Analyses of new input-utterances are constructed out of fragments of the analyses that occur in the corpus. By taking into account the occurrence-frequencies of the fragments, it can be decided which is the most probable analysis that can be constructed in this way.

Following Bod (1995), we note that a specific instance of this data-oriented processing framework can be described by indicating four components:

1.  a definition of a *formal representation for utterance-analyses*

2.  a definition of the *fragments* of the utterance-analyses that may be used as units in constructing an analysis of a new utterance,

3.  a definition of the *operations* that may be used in combining fragments,

4.  a definition of the way in which *the probability of an analysis* of a new utterance is computed on the basis of the occurrence-frequencies of the fragments in the corpus.

The DOP framework thus allows for a wide range of different instantiations. We hypothesize that human language processing can be modelled as a probabilistic process that operates on a corpus of representations of past language experiences, but we leave open how the utterance-analyses in the corpus are represented, what sub-structures of these utterance-analyses play a role in processing new input, and what the details of the probabilistic calculations are.

In this paper we present an in-depth discussion of a data-oriented processing model based on labelled phrase structure trees (Bod, 1992-95). Then we will give an overview of the other models that instantiate the DOP approach. Many of these models also employ labelled phrase-structure trees, but use different criteria for extracting subtrees from the corpus or for computing probabilities (Bod 1996b; Charniak 1996a/b; Goodman 1996; Rajman 1995a/b; Sekine & Grishman 1995; Sima'an 1995-96); other models use richer formalisms for their corpus annotations (van den Berg et al. 1994; Bod et al., 1996a/b; Bonnema 1996; Kaplan 1996; Tugwell 1995).

## 2. A First Data-Oriented Processing System: DOP1

We now define an instance of the DOP framework which we call DOP1. We make specific choices for each of the relevant components identified above. We specify (1) the representations that are assumed, (2) the fragments of these representations that can be used to generate new trees, (3) the operator that is used to combine fragments, and (4) the probabilistic model that is assumed.

### (1) Utterance-analyses

We do not know yet, what kind of representations would provide us with the most suitable stylization of a language user's "syntactic structure experience"; the representations employed by current linguistic theories are plausible candidates to consider for this purpose. But in this paper we will stick to a much simpler system: we will encode utterance-analyses as labelled trees, where the labels are primitive symbols. This notation is obviously limited. It does not represent the meanings of constituents; it ignores "deep" or "functional" syntactic structures that do not coincide with surface structures; and it does not even allow syntactic features such as case, number or gender. We therefore do not expect to stay with this decision for a very long time (see section 6). But for the moment it has two big advantages: it is very simple, and it is the kind of representation that is assumed in readily available annotated corpora such as the Penn Treebank (Marcus et al., 1993).

The representation system is in fact the competence grammar that the system assumes: it defines the set of *possible* analyses. Notice that it is not necessary to define this set very narrowly. Only the linguistic regularities that are exploited by the parsing system or by the probability calculations must be viewed as "hardwired" in the representation system. Other regularities may be expected to emerge as side-effects of the disambiguation process.

Note that a corpus of utterance-analyses is not a set but a "bag" (also known as an "occurrence-set"). Analyses may occur more than once, and how often they occur is significant for the statistical calculations that are to be performed on the corpus.

### (2) Fragments

The fragments of the corpus trees that the system uses as units are *subtrees*. A subtree of a tree *T* is a subgraph *t* of *T* such that

(1) *t* consists of more than one node
(2) *t* is connected
(3) except for the leaf nodes of *t*, each node in *t* has the same daughter-nodes as the corresponding node in *T*

For example, suppose we have the tree *T* shown in Fig. 1 for an utterance of the sentence *John likes Mary*. Then the trees *t* and *u* are valid fragments of *T*, while trees *v* and *w* are not (see Fig. 2).

Given a corpus of trees C, we define the bag of subtrees of C as the bag in which every subtree occurs exactly as often as it can be identified in a tree in C.

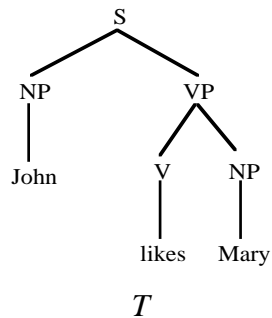

*T*

Figure 1. A corpus analysis tree for *John likes Mary*

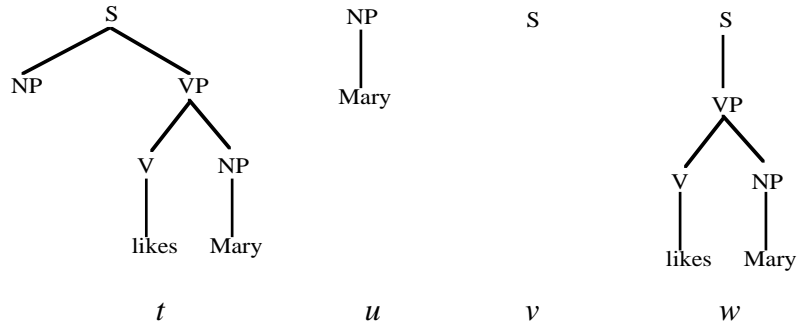

Figure 2. *t* and *u* are subtrees of *T*, but *v* and *w* are not

**(3) Combination operations**

The only combination operation used by DOP1 is the *composition* operation, also called *leftmost substitution* operation. Composition is a partial function on pairs of labelled trees; its range is the set of labelled trees. The composition of tree *t* and tree *u*, written as *t* ∘ *u*, is defined iff the label on the root node of *u* is identical to the label on the leftmost nonterminal leaf node of *t*. If *t* ∘ *u* is defined, it yields a copy of *t* in which a copy of *u* has been substituted on *t*'s leftmost nonterminal leaf node. (The requirement to substitute on the *leftmost* nonterminal makes the composition of two subtrees unique.)

The composition operation is illustrated in Figure 3:

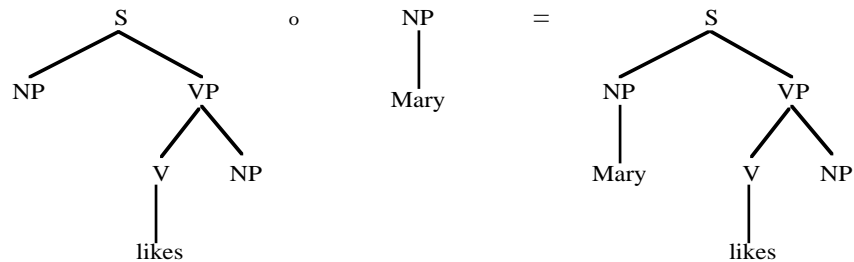

Figure 3. Illustration of the composition operation.

In the sequel we will write (*t* ∘ *u*) ∘ *v* as: *t* ∘ *u* ∘ *v*. Note, however, that composition is not associative.

Given a bag of subtrees B, a sequence of compositions $t_1 \circ \ldots \circ t_n$ with $t_i \in$ B yielding a tree T is called a *derivation* of T.

Given a bag of subtrees B, the set of trees without nonterminal leaves that can be generated by means of iterative composition of members of B is called the *tree-language* generated by B. (This language is a subset of the language defined by the underlying competence model.) The set of strings yielded by these trees is called the *string-language* generated by B.

Given a corpus of trees C, the tree-language generated by the bag of subtrees of the trees of C is said to be the tree-language induced by C. The string-language generated by the bag of subtrees of the trees of C is said to be the string-language induced by C.

**(4) Probability calculation**

By defining a method for extracting subtrees from an annotated corpus and a method for composing such subtrees to construct new trees, we have effectively established a way to view an annotated corpus as a grammar -- i.e., as a definition of a set of strings with syntactic analyses. This grammar becomes a *stochastic* grammar if we redefine the generation of syntactic trees as a stochastic process that takes the frequency-distributions of the corpus-subtrees into account: a process that first chooses a subtree at random among the subtrees with the distinguished root label (for instance S), then composes this subtree with a subtree that is randomly chosen among the ones that it *can* be composed with, and then repeats that step until a tree results without nonterminal leaves. To every tree and every string we assign the probability that it is generated by this stochastic process.

Note that this calculation does *not* use the corpus as a sample for estimating the parameter-values of a stochastic model of a population. The corpus subtrees are used directly as a stochastic generative system, and new input receives the analysis that is most likely to be generated by that system. We thus treat the collection of corpus subtrees as if (1) all subtrees are stochastically independent, and as if (2) the collection contains all subtrees that may be expected to occur in the future. It should be stressed that both assumptions are wrong. The occurrence-probabilities of the subtrees are certainly correlated to some extent, because many distinct overlapping subtrees are derived from one and the same corpus-tree. Assumption (1) may nevertheless be relatively harmless as long as *all* (arbitrarily large) subtrees are taken into account. That assumption (2) is incorrect, becomes most dramatically obvious when an input utterance contains a word that does not appear in any of the corpus-subtrees. This assumption is therefore dropped in Bod (1995, 1996b) where a corpus is treated as a sample of a larger population (see section 5).

Before we go into more detail about the computation of probabilities, let us illustrate the stochastic generation-process by means of an extremely simple imaginary example. Suppose that a corpus of sentence-analyses consists of only the two trees given in Figure 4.

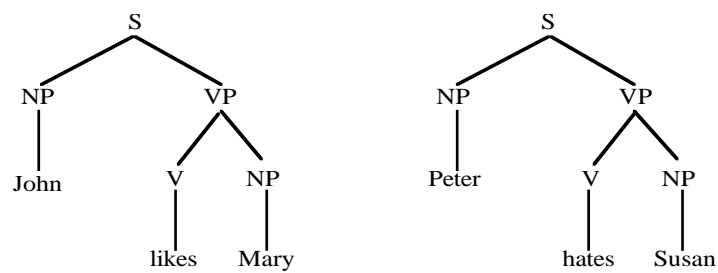

Figure 4. A corpus of two trees

The collection of subtrees extracted from this corpus is represented in Figure 5. (Notice that some subtrees occur twice.)

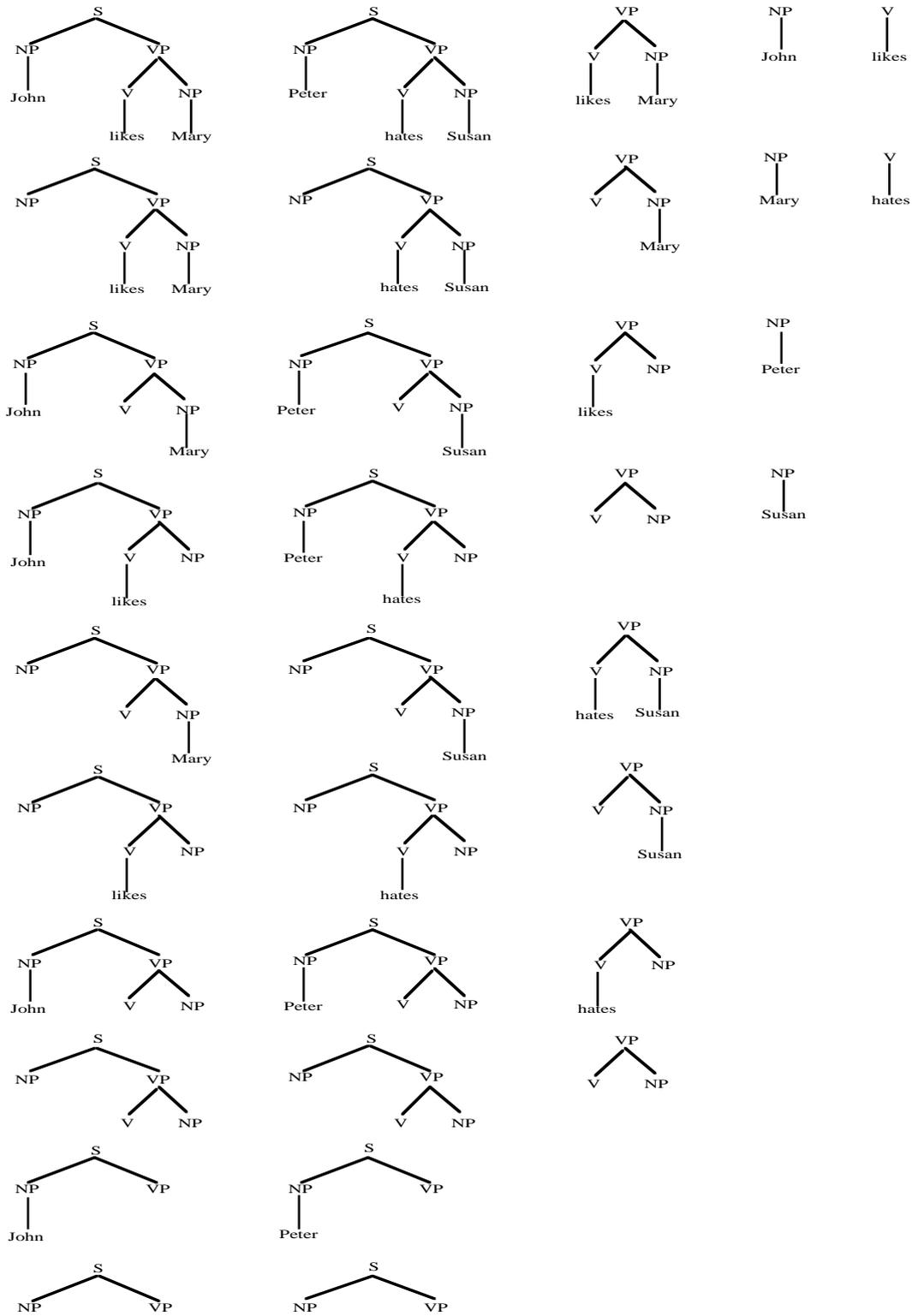

Figure 5. The collection of subtrees corresponding to the corpus in Figure 4.

By means of the composition operation, new sentence-analyses can be constructed out of this subtree collection. For instance:

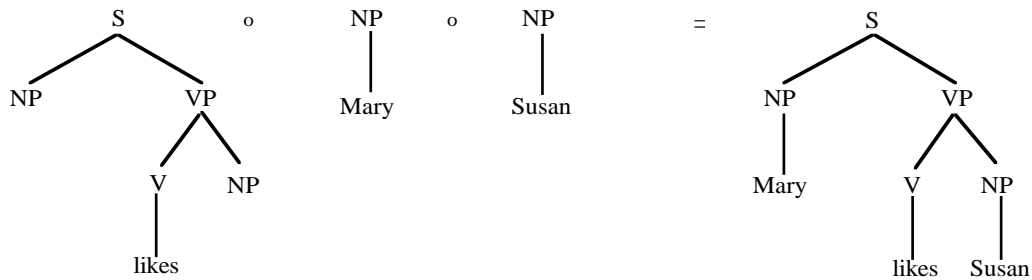

Figure 6. Generating "*Mary likes Susan*" by combining subtrees.

The probability of this particular derivation is the joint probability of 3 stochastic events:

(1) selecting the subtree $_S$[NP $_{VP}$[$_V$[likes] NP]] among the subtrees with root label S,

(2) selecting the subtree $_{NP}$[Mary] among the subtrees with root label NP,

(3) selecting the subtree $_{NP}$[Susan] among the subtrees with root label NP.

Since the events are assumed to be stochastically independent, their joint probability is the product of the probabilities of the individual events. Thus, the probability of the derivation of Figure 6 is:

*P*(*t* = $_S$[NP $_{VP}$[$_V$[likes] NP]] | *root*(*t*) = S) * *P*(*t* = $_{NP}$[Mary] | *root*(*t*) = NP) * *P*(*t* = $_{NP}$[Susan] | *root*(*t*) = NP)

We assume the selection of subtrees from the collection to be a random process. Conditional probabilities can therefore be computed by dividing the cardinalities of the occurrences of the relevant kinds of trees. For instance:

*P*(*t* = $_S$[NP $_{VP}$[$_V$[likes] NP]] | *root*(*t*) = S) = #($_S$[NP $_{VP}$[$_V$[likes] NP]]) / #(*t* | *root*(*t*) = S)

where #(*x*) denotes the number of occurrences of the subtree-type *x*. The probability of the derivation of Figure 6 is therefore:

#($_S$[NP $_{VP}$[$_V$[likes] NP]]) / #(*t* | *root*(*t*) = S) * #($_{NP}$[Mary]) / #(*t* | *root*(*t*) = NP) * #($_{NP}$[Susan]) / #(*t* | *root*(*t*) = NP) =

1/20 * 1/4 * 1/4 =

1/320

In general, a derivation $t_1 \circ ... \circ t_n$ thus has a probability

$$P(t_1 \circ ... \circ t_n) = \Pi_i \#(t_i) / \#(t \mid root\ (t) = root\ (t_i))$$

The probability of a parse tree is computed by considering all its derivations. For instance, the parse tree derived in Figure 6 may also be derived as in Figure 7 or Figure 8.

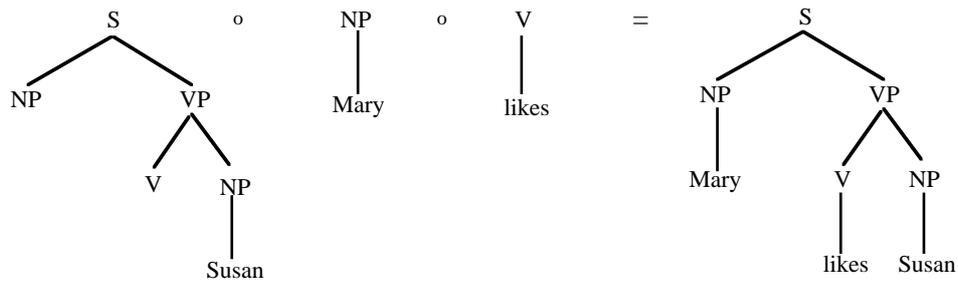

Figure 7. A different derivation, yielding the same parse tree.

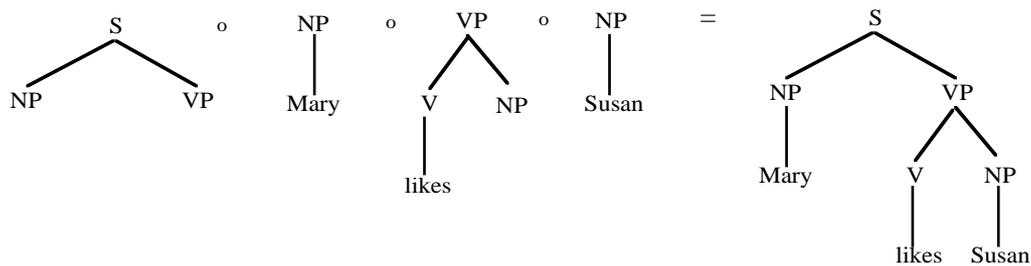

Figure 8. One more derivation yielding the same parse tree.

Thus, a parse tree can be generated by a large number of different derivations, that involve different fragments from the corpus. Each of these derivations has its own probability of being generated. For instance, the following shows the probabilities of the three example derivations given above.

| | | | |
|---|---|---|---|
| $P$(Figure 6) | = | 1/20 * 1/4 * 1/4 | = 1/320 |
| $P$(Figure 7) | = | 1/20 * 1/4 * 1/2 | = 1/160 |
| $P$(Figure 8) | = | 2/20 * 1/4 * 1/8 * 1/4 | = 1/1280 |

Table 1. Probabilities of the derivations of Figures 6, 7 and 8.

The probability that a parse is generated equals the probability that any of its distinct derivations is generated. Therefore, the probability of a parse is the sum of the probabilities of all its derivations. If a tree has $k$ distinct derivations, $(t_{11} \circ ... \circ t_{1j} \circ ...)$, ..., $(t_{i1} \circ ... \circ t_{ij} \circ ...)$, ..., $(t_{k1} \circ ... \circ t_{kj} \circ ...)$, its probability can be written as: $\Sigma_i \ \Pi_j \ \#(t_{ij}) / \#(t \mid root \ (t) = root \ (t_{ij}) )$.

We note that sentences may be ambiguous, i.e., they may have different parse trees. The probability of generating a sentence is therefore the sum of the probabilities of its parses.

Formally, DOP1 can be viewed as a Stochastic Tree-Substitution Grammar (STSG). It is useful, therefore, to introduce this formalism.

### Stochastic Tree-Substitution Grammars

A *Stochastic Tree-Substitution Grammar G* is a five-tuple $<V_N, V_T, S, R, P>$ where $V_N$ is a finite set of nonterminal symbols and $V_T$ is a finite set of terminal symbols. $S \in V_N$ is the distinguished nonterminal symbol. $R$ is a finite set of elementary trees whose top nodes and interior nodes are labeled by nonterminal symbols and whose yield nodes are labeled by terminal or nonterminal symbols. $P$ is a function which assigns to every elementary tree $t \in R$ a probability $P(t)$. For a tree $t$ with a root $\alpha$, $P(t)$ is interpreted as the probability of substituting $t$ on $\alpha$. We require, therefore, that $0 < P(t) \leq 1$ and $\Sigma_{t:root(t) = \alpha} P(t) = 1$.

If $t_1$ and $t_2$ are trees such that the *leftmost nonterminal yield node* of $t_1$ is equal to the *root* of $t_2$, then $t_1 \circ t_2$ is the tree that results from substituting $t_2$ on this leftmost nonterminal yield node in $t_1$. The partial function $\circ$ is called *leftmost substitution* or simply *substitution*. We write $(t_1 \circ t_2) \circ t_3$ as $t_1 \circ t_2 \circ t_3$, and in general $(..((t_1 \circ t_2) \circ t_3) \circ ..) \circ t_n$ as $t_1 \circ t_2 \circ t_3 ... \circ t_n$.

A *leftmost derivation* generated by an STSG $G$ is a tuple of trees $<t_1,...,t_n>$ such that $t_1,...,t_n$ are elements of $R$, the root of $t_1$ is labeled by $S$ and the yield of $t_1 \circ ... \circ t_n$ is labeled by terminal symbols. The set of leftmost derivations generated by $G$ is thus given by

$$Derivations(G) = \{<t_1,...,t_n> \mid t_1,...,t_n \in R \ \wedge \ root(t_1) = S \ \wedge \ yield(t_1 \circ ... \circ t_n) \in V_T^+\}$$

For convenience we will use the term derivation for leftmost derivation. A derivation $<t_1,...,t_n>$ is called a derivation of tree $T$, iff $t_1 \circ ... \circ t_n = T$. A derivation $<t_1,...,t_n>$ is called a derivation of string $s$, iff $yield(t_1 \circ ... \circ t_n) = s$. The probability of a derivation $<t_1,...,t_n>$ is defined as $P(t_1) * ... * P(t_n)$.

A *parse tree* generated by an STSG $G$ is a tree $T$ such that there is a derivation $<t_1,...,t_n> \in Derivations(G)$ for which $t_1 \circ ... \circ t_n = T$. The set of parse trees, or *tree language*, generated by $G$ is given by

$$Parses(G) = \{T \mid \exists <t_1,...,t_n> \in Derivations(G) : t_1 \circ ... \circ t_n = T\}$$

For reasons of conciseness we will often use the terms *parse* or *tree* for a parse tree. A parse whose yield is equal to string $s$, is called a parse of $s$. The probability of a parse is defined as the sum of the probabilities of all its derivations.

A *string* generated by an STSG *G* is an element of $V_T{}^+$ such that there is a parse generated by *G* whose yield is equal to the string. The set of strings, or *string language*, generated by *G* is given by

$$Strings(G) = \{s \mid \exists\, T : T \in Parses(G) \;\wedge\; s = yield(T)\}$$

The probability of a string is defined as the sum of the probabilities of all its parses. The probability of a string thus also equals the sum of the probabilities of all its derivations.

It is evident that DOP1 is an instantiation of STSG. DOP1 projects a corpus of tree structures into an STSG, where the subtrees of DOP1 are the elementary trees of the STSG, and where the corpus probabilities of the subtrees of DOP1 are the probabilities of the corresponding elementary trees of the STSG.

Bod (1995) studied the question of *stochastic strong generative power* of STSG: given an STSG, can other formalisms be used to formulate stochastic grammars that generate the same trees with the same probabilities? Among other things, it is shown there that STSG is stochastically stronger than stochastic context-free grammar (Booth, 1969; Suppes, 1970) and stochastic history-based grammar (Smith, 1973; Black et al., 1993).

## 3.    Parsing and disambiguation with DOP1

We consider the problem of computing the most probable parse tree of an input utterance, as defined by DOP1. The algorithm we discuss does not exploit the particular properties of Data-Oriented Processing; it works with any Stochastic Tree-Substitution Grammar. We distinguish between parsing and disambiguation. By parsing we mean generating a parse forest for an input sentence. By disambiguation we mean the selection of the most probable parse from the forest.

### 3.1 Parsing

The algorithm that creates a parse forest for an input sentence is derived from algorithms that exist for Context-Free Grammars, which parse an input sentence of *n* words in polynomial (usually in cubic) time. These parsers make use of a chart or well-formed substring table. They take as input a set of context-free rewrite rules and a sentence and produce as output a chart of labeled phrases. A labeled phrase is a sequence of words labeled with a category symbol which denotes the syntactic category of that phrase. A chart-like parse forest can be obtained by including pointers from a category to the other categories which caused it to be placed in the chart. Algorithms that accomplish this can be found in e.g. (Kay, 1980; Winograd, 1983; Jelinek et al., 1990).

The chart parsing approach can be applied to parsing with Stochastic Tree-Substitution Grammars if we note that every elementary tree *t* of the STSG can be viewed as a context-free rewrite rule: *root*(*t*) → *yield*(*t*). In order to obtain a chart-like forest for a sentence parsed with an STSG, we label the phrases not only with their syntactic categories but with their full elementary trees. Note that in a chart-like forest generated by an STSG, different derivations that generate identical trees do not collapse. We will therefore talk about a *derivation forest* generated by an STSG (cf. Sima'an et al., 1994). The following formal example illustrates what such a derivation forest may look like. In the example, we leave out the probabilities of the elementary trees, that will be needed in the disambiguation phase. The visual representation is based on (Kay, 1980): every entry ($i,j$) in the chart is indicated by an edge and spans the words between the $i$-th and the $j$-th position of a sentence. Every edge is labeled with linked elementary trees that constitute subderivations of the underlying subsentence. The set of elementary trees of the example STSG consists of the trees in Figure 9.

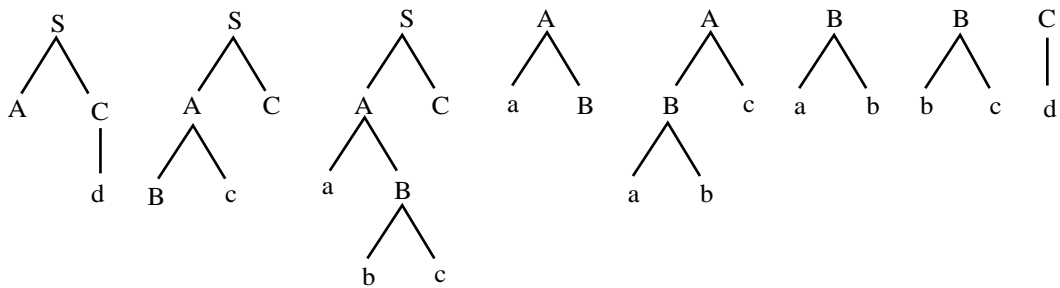

Figure 9. Elementary trees of an example STSG.

For the input string **abcd**, this gives the derivation forest in Figure 10. Note that some of the derivations in the forest generate the same tree. By exhaustively unpacking the forest, four different derivations generating two different trees are obtained. Both trees are generated twice, by different derivations (with possibly different probabilities). We may ask whether we can pack the forest by collapsing spurious derivations, summing up their probabilities. Unfortunately, no efficient procedure is known that accomplishes this. (Remember that there can be exponentially many derivations for one tree).

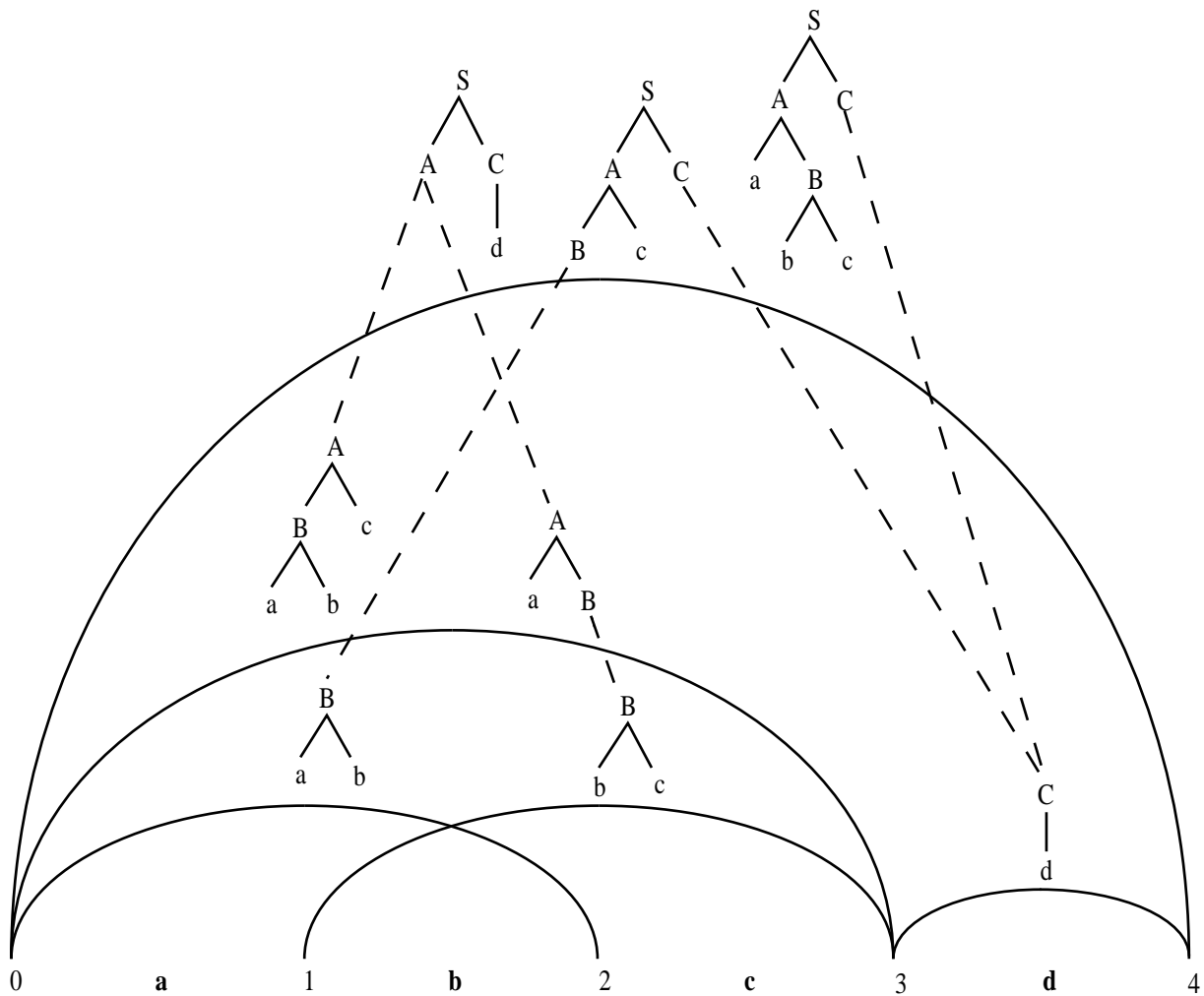

Figure 10. Derivation forest for the string **abcd**.

## 3.2 Disambiguation

Polynomial time parsing does not guarantee polynomial time disambiguation. A sentence may have exponentially many parses and any such parse may have exponentially many derivations. Therefore, in order to find the most probable parse of a sentence, it is not efficient to compare the probabilities of the parses by exhaustively unpacking the chart. Even for determining the probability of one parse, it is not efficient to add the probabilities of all derivations of that parse.

**Viterbi optimization is not applicable to finding the most probable parse**

When parsing on the basis of an SCFG (stochastic context-free grammar) or an STSG, the most probable derivation of a sentence can be selected in cubic time by means of a strategy known as the Viterbi algorithm (Viterbi, 1967; Fujisaki et al., 1989; Jelinek et al., 1990). When an SCFG is used, the most probable derivation also generates the most probable parse (since every parse is generated by exactly one derivation). But when an STSG is used, this is not the case -- since the probability of a parse is the sum of the probabilities of all its different derivations.

Let us look in more detail at the Viterbi algorithm. The basic idea of Viterbi is the elimination of low probability subderivations in a bottom-up fashion. Two different subderivations of the same part of the sentence whose resulting subparses have the same root can both be developed (if at all) to derivations of the whole sentence in the same ways. Therefore, if one of these two subderivations has a lower probability, it can be eliminated. This is illustrated by a formal example in Figure 11. Suppose that during bottom-up parsing of the string *abcd* two subderivations *d1* and *d2* have been generated for the substring *abc*, with the following resulting subparses.

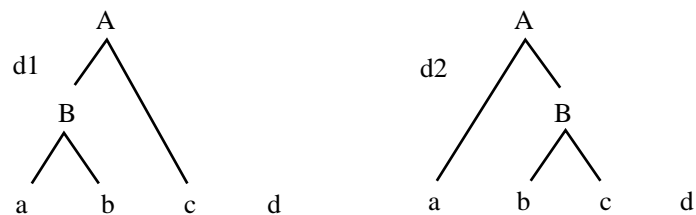

Figure 11. Two subderivations *d1* and *d2* for the substring *abc*.

If the probability of *d1* is higher than the probability of *d2*, we can eliminate *d2* if we are only interested in finding the most probable <u>derivation</u> of *abcd*. But if we are interested in finding the most probable <u>parse</u> of *abcd* (generated by an STSG), we are not allowed to eliminate *d2*. This can be seen by the following. Suppose that we have the additional elementary tree given in Figure 12.

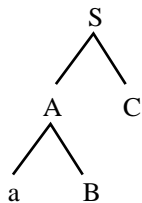

Figure 12. Additional elementary tree.

This elementary tree may be developed to a tree that can also be developed by *d2*, but not to a tree that can be developed by *d1*. And since the probability of a parse tree is equal to the sum of the probabilities of all its derivations, it is still possible that *d2* contributes to the generation of the most probable parse. Therefore we are not allowed to eliminate *d2*.

This counter-example does not prove that there is no optimization that allows for polynomial time selection of the most probable parse. But it makes clear that a *best-first* search, as accomplished by Viterbi, is not adequate for finding the most probable parse in STSG. In the past few years, several researchers have tried to find optimizations. But the algorithms that were found either turned out to be still exponential (Sima'an et al., 1994), or did not refer to the most probable parse (Goodman, 1996; see section 5). Finally, Sima'an (1996b) proved that there is no deterministic polynomial time algorithm for finding the most probable parse of a sentence in STSG (i.e., the problem of computing the most probable parse in STSG is NP-hard).

We should of course ask how often the most probable derivation in fact happens to yield the most probable parse. The experimental results we discuss in section 4, however, indicate that calculating the most probable parse is indeed more accurate than choosing the parse generated by the most probable derivation.

## Monte Carlo disambiguation: estimating the most probable parse by sampling random derivations

Although there is no deterministic polynomial algorithm for finding the most probable parse, it may still be that there is an algorithm that *estimates* a most probable parse with an error that can be made arbitrarily small. In this section, we will deal with this question.

We have seen that a best-first search, as accomplished by Viterbi, can be used for finding the most probable derivation in STSG but not for finding the most probable parse. If we apply instead of a best-first search, a *random-first* search, we can generate a random derivation from the derivation forest -- provided that the random choices are based on the actual probabilities of the subderivations. By iteratively generating a large number of random derivations we can estimate the most probable parse as the parse which results most often from these random derivations (since the probability of a parse is the probability that any of its derivations occurs). The most probable parse can be estimated as accurately as desired by making the number of random samples sufficiently large. According to the Law of Large Numbers, the most often generated parse converges to the most probable parse. Methods that estimate the probability of an event by taking random samples are known as Monte Carlo methods (Meyer, 1956; Hammersley & Handscomb, 1964).

The selection of a random derivation can be accomplished in a bottom-up fashion analogous to Viterbi. Instead of selecting the most probable subderivation at each node-sharing in the chart, a random subderivation is selected at each node-sharing (in such a way that a subderivation that has $n$ times as large a probability as another subderivation also has $n$ times as large a chance to be chosen as this other subderivation). Once arrived at the S-node, the random derivation of the whole sentence can be retrieved by tracing back the choices made at each node-sharing. We may of course postpone sampling until the S-node, such that we sample directly from the distribution of all S-derivations. But this would take exponential time, since there may be exponentially many derivations for the whole sentence. By sampling bottom-up at every node where ambiguity appears, the maximum number of different subderivations at each node-sharing is bounded to a constant (the total number of rules of that node), and therefore the time complexity of generating a random derivation of an input sentence is equal to the time complexity of finding the most probable derivation. This is exemplified by the following algorithm.

Algorithm 1: Sampling a random derivation from a derivation forest

Given a derivation forest of a sentence of $n$ words, consisting of labeled entries $(i,j)$ that span the words between the $i$-th and the $j$-th position of the sentence. Every entry is labeled with linked elementary trees, together with their probabilities, that constitute subderivations of the underlying subsentence. Sampling a derivation from the chart consists of choosing at every labeled entry (bottom-up, breadth-first)[1] at random a subderivation of each root-node:

for $k$ := 1 to $n$ do
   for $i$ := 0 to $n$-$k$ do
     for chart-entry $(i,i+k)$ do
       for each root-node $X$ do
         select at random a subderivation of root-node $X$
         eliminate the other subderivations

Let { $(e_1, p_1)$ , $(e_2, p_2)$ , ... , $(e_n, p_n)$ } be a probability distribution of events $e_1$, $e_2$, ..., $e_n$; an event $e_i$ is said to be *randomly selected* iff its chance of being selected is equal to $p_i$. In order to allow for "simple sampling" (Cochran, 1963), where every event has an equal chance of being selected, one must convert a probability distribution into a sample space such that the frequency of occurrence $r_i$ of each event $e_i$ is a positive integer equal to $n{\cdot}p_i$, where $n$ is the size of the sample space.

We thus have an algorithm that selects a random derivation from a derivation forest. The parse tree that results from this derivation constitutes a first guess for the most probable parse. A more reliable guess can be computed by sampling a larger number of random derivations, and selecting the parse which results most often from these derivations. How large should this sample set be chosen?

Let us first consider the probability of error: the probability that the parse that is most frequently generated by the sampled derivations, is in fact not equal to the most probable parse. An upper bound for this probability is given by $\sum_{i \neq 0} (1 - (\sqrt{p_0} - \sqrt{p_i})^2)^N$, where the different values of $i$ are indices corresponding to the different parses, $0$ is the index of the most probable parse, $p_i$ is the probability of parse $i$; and $N$ is the number of derivations that was sampled. (Cf. Hammersley & Handscomb, 1964; Deming, 1966.)

This upper bound on the probability of error becomes small if we increase $N$, but if there is an $i$ with $p_i$ close to $p_0$ (i.e., if there are different parses in the top of the sampling distribution that are almost equally likely), we must make $N$ very large to achieve this effect. If there is no unique most probable parse, the sampling process will of course not converge on one outcome.

---

[1] The heuristics of sampling bottom-up, breadth-first can be changed into any other order (e.g.: top-down, depth-first). We chose the current algorithm for its analogy with Viterbi (cf. Jelinek et al., 1990).

In that case, we are interested in all of the parses that outrank all the other ones. But also when the probabilities of the most likely parses are very close together without being exactly equal, it is not *the* most probable parse we are interested in, but the set of all these almost equally highly probable parses. In that case there is an ambiguity which cannot be resolved by probabilistic syntactic considerations; the syntactic disambiguation component of a language processing system should recognize this fact and keep the relevant set of alternative analyses for further consideration by other system components.

We must conclude, therefore, that it is misleading to construe the task of the syntactic disambiguation component as the calculation of the most probable parse. The task is the calculation of the probability distribution of the various possible parses. When we estimate this probability distribution by statistical methods, we must establish the reliability of this estimate. This reliability is characterized by the probability of significant errors in the estimates of the probabilities of the various parses.

If a parse has probability $p_i$, and we try to estimate the probability of this parse by its frequency in a sequence of $N$ independent samples, the variance in the estimated probability is $p_i(1 - p_i)/N$. Since $0 < p_i \leq 1$, the variance is always smaller than or equal to $1/(4N)$. Thus, the standard error $\sigma$, which is the square root of the variance, is always smaller than or equal to $1/(2\sqrt{N})$. This allows us to calculate a lower bound for $N$ given an upper bound for $\sigma$, by $N \geq 1/(4\sigma^2)$. For instance, we obtain a standard error $\leq 0.05$ if $N \geq 100$.

We thus arrive at the following algorithm:

Algorithm 2: Estimating the parse probabilities.

Given a derivation forest of a sentence and a threshold $\sigma_M$ for the standard error:

$N$ := the smallest integer larger than $1/(4\sigma_M^2)$
repeat $N$ times:
    sample a random derivation from the derivation forest
    store the parse generated by this derivation
for every parse $i$ :
    $p_i$ := #($i$) / $N$

In the case of a forced choice experiment we choose the parse with the highest probability from this distribution. Rajman (1995a) gives a correctness proof for Monte Carlo disambiguation.

## 4.    Experimental properties of DOP1

Some properties of DOP1 were established by experiments on a set of 600 trees from the Penn Treebank ATIS corpus (Marcus et al., 1993). All trees were stripped of their words, but were not edited any further. The resulting corpus of 600 trees of part-of-speech sequences was divided at random into 500 training set trees and 100 test set trees. The training set trees were converted into subtrees, yielding roughly 350 000 distinct subtrees, and were enriched with their corpus probabilities. The part-of-speech sequences from the test set served as test sentences that were parsed using the subtrees from the training set (for the parsing of word strings see section 5). The most probable parses were estimated from probability distributions of 100 sampled derivations.

We used the following *accuracy metrics* from the literature to compare the most probable parses generated by the system with the test set parses: (1) the *parse accuracy*, defined as the percentage of the selected parses that are identical to the corresponding test set parse (Magerman & Marcus, 1991; Magerman, 1994); (2) the *sentence accuracy*, defined as the percentage of the selected parses in which no brackets cross the brackets in the corresponding test set parse (Brill, 1993; Schabes et al., 1993); and (3) the *bracketing accuracy*, the percentage of the brackets of the selected parses that do not cross the brackets in the corresponding test set parse (Black et al., 1991; Pereira & Schabes, 1992). For measuring the sentence accuracy and the bracketing accuracy, the output parses were maded binary branching, since otherwise these measures are meaningless (cf. Magerman, 1994).

It may be clear that the parse accuracy is the most stringent metric, while the bracketing accuracy is the weakest metric. Remember that in testing a performance model, we are first of all interested in whether the model can correctly select the perceived parse from the possible parses of a sentence. Therefore, the parse accuracy metric is the most significant one.

**(1) Accuracy increases if larger corpus fragments are used**

DOP1 takes arbitrarily large subtrees into consideration to estimate the probability of a parse. In order to test the usefulness of this feature, we performed experiments with versions of DOP1 where the subtree collection is restricted to subtrees with a certain maximum depth (where the depth of a tree is defined as the length of the longest path from the root to a leaf). The following table shows the results of these experiments.

| depth of subtrees | parse accuracy | sentence accuracy | bracketing accuracy |
|:---:|:---:|:---:|:---:|
| 1 | 27% | 59% | 88.1% |
| ≤2 | 45% | 65% | 90.2% |
| ≤3 | 56% | 73% | 91.9% |
| ≤4 | 59% | 73% | 91.8% |
| ≤5 | 61% | 74% | 93.2% |
| ≤6 | 62% | 75% | 93.5% |
| unbounded | 64% | 75% | 94.8% |

Table 2. Accuracies against subtree-depth for ATIS.

The table shows a considerable increase in accuracy, for all metrics, when enlarging the maximum depth of the subtrees from 1 to 2 (note that  for depth 1, DOP1 is equivalent to a stochastic context-free grammar). The accuracies keep increasing, at a slower rate, when the depth is enlarged further. The highest accuracies are achieved by using all subtrees from the training set: 64% parse accuracy, 75% sentence accuracy and 94.8% bracketing accuracy. As expected, the bracketing accuracy is higher than the sentence accuracy which is higher than the parse accuracy. The *coverage* of the system (defined as the percentage of the test sentences for which a parse was found) was 98%.

A parse accuracy of 64% seems a less than exciting result. But we should keep in mind that the annotations in the Penn Treebank contain many inconsistencies (cf. Ratnaparkhi, 1996). A single inconsistency in a test set tree will very likely yield a zero percent parse accuracy for the particular test set sentence. We have therefore also conducted experiments with ATIS data that were manually cleaned up, resulting in much higher parse accuracies, up to 96% (cf. Bod 1995, 1996b). However, the accuracies displayed the same kind of improvement if larger subtrees were used.

**(2)  The most probable derivation is less accurate than the most probable parse.**

DOP1 disambiguates by selecting the most probable parse rather than the most probable derivation. In section 3.2 we saw that this is computationally unattractive, since it precludes the use of Viterbi optmization. One may wonder, therefore, whether the most probable derivation may in fact be a better disambiguation criterion, or whether the two criteria are perhaps closely correlated.

To study this, we calculated the accuracies for a version of the system which selects the trees generated by the most probable derivations. For the same training/test set split, the parse accuracy decreased from 64% to 46%, the sentence accuracy from 75% to 68%, and the bracketing accuracy from 94.8% to 92.0%. (These are the results if subtrees of unlimited depth are taken into account; if we limit the subtree depth, accuracies decrease in a similar way as in Table 2.) We conclude that predictions based on the most probable derivation are considerably less accurate than predictions based on the most probable parse.

An interesting result by Sima'an (1996a) should, however, be noted in this connection. He reports on an algorithm computing the most probable derivation which obtains practically the same results as the most probable parse algorithm on the ATIS corpus. This algorithm uses a different set of corpus subtrees than DOP1: it only uses subtrees with at most two non-terminal leaf nodes (see section 5).

**(3)  Accuracy decreases if once-occurring fragments are ignored**

There is an important question as to whether we need *all* corpus fragments for accurately predicting the perceived parse of a string, or whether we can safely throw away the once-occurring fragments (so-called *hapaxes*). After having eliminated all once-occurring subtrees from the training set, all accuracies decreased: the parse accuracy from 64% to 60%, the sentence accuracy from 75% to 70%, and the bracketing accuracy from 94.8% to 93.3%. Distinguishing between small and large hapaxes, showed that the accuracies were practically not affected by eliminating the hapaxes of depth 1. Eliminating hapaxes larger than depth 1, decreased the accuracies. We conclude that large subtrees, even if they occur once, contribute to the prediction of the correct parse.

## 5.    Alternative Data-Oriented Processing Models

The model discussed above closely corresponds to the first fully articulated DOP model, presented in Bod (1992). Several other instantiations of this approach have been proposed in the meantime. Many of these assume the same annotation formalism as DOP1, but try to improve the efficiency and/or to extend the model in order to cope with unknown words. They explore different criteria for extracting subtree collections from an annnotated corpus, and/or employ different disambiguation strategies. We briefly discuss these proposals in the current section. (Other models have explored the possibilities of using richer corpus annotation formalisms; these we discuss in the next section.)

### Using unknown fragments for coping with unknown words: Bod (1995-96b)

Bod (1995, 1996b) studies how DOP1 can be extended to parse word strings. He shows that the problem of word string parsing does not only lie in unknown words, but also in ambiguous words that occur in the training set with a different lexical category than is needed for parsing the test sentence. Bod calls these words "unknown-category words". He argues that in order to cope with unknown and unknown-category words, a data-oriented processing system must compute the probabilities of *unknown subtrees*. An unknown subtree is a subtree which does not occur in the training set, but which may show up in an additional sample of trees.

   To derive unknown subtrees, Bod allows unknown (-category) words of a sentence to mismatch with the lexical items of the training set subtrees. The Good-Turing method (Good, 1953) is used to estimate the probabilities of unknown subtrees. In his experiments, Bod restricts the mismatches to unknown words and a restricted set of potentially unknown-category words. On ATIS word strings, he achieves very competitive results, without the need of an external part-of-speech tagger.

### Using the most likely derivation: Sima'an (1995, 1996a)

Sima'an (1995, 1996a) gives an efficient Viterbi-style deterministic algorithm for selecting the parse corresponding to the most probable derivation of a sentence in DOP1 and STSG. His algorithm initially uses the CFG-backbone of an STSG for restricting the parse space of a sentence, after which it imposes the contraints embedded by the STSG to further restrict the parse space and to select the "best analysis".

Applying his algorithm to an STSG defined by a collection of subtrees extracted from an annotated corpus, Sima'an reports on experiments with different maximal subtree depths that display a similar improvement in accuracy as our experiments in section 4. Besides varying the subtree depths, Sima'an also varies the number of terminals and nonterminals ("substitution sites") in the frontiers of the subtrees. He shows that the accuracy of the most likely derivation increases if the subtrees are restricted to those with maximally two nonterminals in the frontiers. Applying this method to the ATIS corpus, he obtains practically the same results as the most probable parse method.

### Using only `S` and `NP` fragments: Sekine & Grishman (1995)

Sekine and Grishman (1995) give an instantiation of DOP1 which constrains the corpus fragments to subtrees rooted with `S` and `NP`. They convert the `S` and `NP` fragments into rules and apply the same probability calculation as DOP1, that is, they take the relative frequencies of these rules and normalize them by their lefthand nonterminal. So far, the model is an instantiation of DOP1, and can be used for parsing part-of-speech strings. To allow for word parsing, Sekine and Grishman interface their model with part-of-speech probabilities in a way which is different from Bod (1995, 1996b) and which looses the statistical dependencies between words and syntactic structure. However, Sekine and Grishman show that their system can rather efficiently and accurately parse word strings from the Penn Treebank Wall Street Journal corpus.

The main reason for restricting the fragments to `S` and `NP` is efficiency. However, their restriction puts an upper bound on coverage and accuracy. In a recent personal communication, Sekine reports that by allowing 5 different nonterminals instead of 2, the accuracy increases considerably.

### Using only fragments of depth one: Charniak (1996a/b)

In Charniak (1996a/b), a model is given equivalent to an instantiation of DOP1 in which the subtrees are constrained to depth one. A subtree of depth one uniquely corresponds to a context-free grammar rule, and as mentioned in section 4, such a DOP model is formally equivalent to a stochastic context-free grammar (SCFG). Charniak uses the same probability calculations as DOP1, that is, the probability of each rule $r$ is computed as the number of appearances of that rule in the treebank divided by the number of appearances of rules with the same lefthand nonterminal as $r$. He then applies a best-first parsing algorithm for computing the most probable parse (which for subtrees of depth one is equivalent to the most probable derivation).

Charniak tested this model on part-of-speech strings from Penn's Wall Street Journal (WSJ) corpus, and shows that it "outperforms all other non-word-based statistical parsers/grammars on this corpus". Thus, even if Charniak's approach constitutes a less-than-optimal version of DOP1, it already outperforms the more sophisticated training and learning approaches such as Inside-Outside training (Pereira & Schabes, 1992; Schabes et al., 1993) and Transformation-Based learning (Brill, 1993). We conjecture that Charniak's results will be even better if also larger fragments are taken into account.

**Using the Maximum Constituents Parse: Goodman (1996)**

Goodman (1996) gives a polynomial parsing strategy for an interesting variation of DOP1. Instead of computing the most probable derivation or parse of a sentence, Goodman gives an algorithm which computes the so-called "maximum constituents parse". The maximum constituents parse of a sentence is the parse tree which maximizes the expected number of correct constituents. Goodman shows that the maximum constituents parse can produce trees that cannot be produced by the DOP1 model, which means that Goodman's model is not an instantiation of DOP1.[2] It does, however, share many features with DOP1: it uses the same subtrees with the same probabilities, and the maximum constituents parse is computed by summing over all derivations.

In his experiments on ATIS, Goodman shows that his method obtains better results than a grammar re-estimation approach based on the Inside-Outside training algorithm (Pereira & Schabes, 1992; Schabes et al., 1993). He performs a statistical analysis using *t*-test on the differences between his results and the results of the Pereira & Schabes method, showing that the differences are statistically significant for a cleaned-up ATIS corpus. The approach of directly using untrained corpus-fragments thus outperforms an approach based on training techniques.

**6. Data-oriented processing models using richer corpus representations**

In the following, we give a summary of the DOP models that assume corpus representations that are different from DOP1.

**A DOP model for Logical-Semantic representations: van den Berg et al. (1994)**

If one wants to use DOP not only as a model for language processing but also for language understanding, it is necessary to develop an instantiation that can deal with semantically analyzed corpus utterances. Van den Berg, Bod and Scha (1994) propose such a model in which the syntactic corpus trees of DOP1 are enriched with semantic annotations by adding to every meaningful syntactic node a type-logical formula that expresses the meaning of the corresponding constituent. As in the purely syntactic versions of DOP, the probability of a (semantic) analysis is computed by considering all the different ways in which it can be generated by combining subtrees from the corpus.

However, the algorithm needed for extracting a subtree out of a tree requires a procedure which can inspect the semantic formula of a node and determine the contribution of the semantics of a lower node, in order to "factor out" that contribution. The details of this procedure have not been specified. One can avoid the need for this procedure by using an annotation which indicates explicitly how the semantic formula for a node is built up on the basis of the semantic formulas of its daugther nodes.

---

[2] See Bod (1996a) for further details.

In Bonnema (1996) and Bod, Bonnema & Scha (1996a), the semantic annotation of the corpus trees is therefore indicated as follows. (1) For every meaningful lexical node a type-logical formula is specified that represents its meaning. (2) For every meaningful non-lexical node a formula schema is specified which indicates how its meaning representation may be put together out of the formulas assigned to its daughter nodes. This representation convention obviously assumes that the meaning representation of a surface-constituent *can* in fact always be composed out of the meaning representations of its subconstituents. This assumption is not unproblematic; however, as a first computational approach to data-oriented semantics it turned out to be powerful enough to annotate almost all ATIS sentences (Bonnema, 1996). The only further novelty of this DOP model is a slight modification in the operation which combines subtrees. If a subtree is extracted out of a tree, the semantics of the new leaf node is replaced by a unification variable of the same type. Correspondingly, when the composition operation substitutes a subtree at this node, this unification variable is unified with the semantic formula on the substituting tree.

The experiments performed by Bonnema (1996), using the parser developed by Sima'an (1995), show that 88% of the ATIS test sentences obtain the correct semantic interpretation, while only 62% obtain a fully correct syntactic structure (see also Bod, Bonnema & Scha, 1996a). Thus, semantic parsing is very robust, even if the underlying syntactic structure does not fully match the syntactic structure of the test set analysis. The semantic annotations also contribute to better time performance: *with* semantic annotations the test sentences are parsed 6 times as fast as *without* semantic annotations.

## A DOP model for State-Transition representations: Tugwell (1995)

Tugwell (1995) raises the fundamental question as to what kind of representation-formalism is most suited to a DOP approach that will do justice to both statistical and structural aspects of language. He sets out a number of requirements which he thinks such a formalism should satisfy: it must have a close correspondence to a Markov process, it should be radically lexicalist, it should be capable of capturing fully the linguistic intuitions of language users, and it should assign an analysis to every interpretable utterance. On the basis of these requirements, Tugwell develops a formalism in the vein of dependency syntax which he calls State-Transition Grammar. The formalism's backbone is a finite-state grammar in which each state-transition (associated with an individual word) can be labelled by a stack of categories. Such a formalism is equivalent to indexed grammar (Aho, 1968), and if restricted to right linear trees, it becomes equivalent to context-free grammar.

Tugwell then defines a DOP model by assuming a corpus analyzed according to this representation formalism. For parsing fresh text with such a corpus, Tugwell collects for each word type the transitions with which it occurs, and calculates the probability of a transition as its number of occurrences divided by the number of all other transitions of the same word type. To find the most probable parse for a sentence, Tugwell simply computes the path from word to word which maximizes the product of the state transitions. This means that Tugwell only takes the minimal fragments of the corpus into account (the single transitions), effectively leading to a first order Markov process. In order to cope with sparse transitions, Tugwell proposes to generalize over a number of transition-features.

Tugwell accomplished a preliminary experiment with a small part of the Brown corpus (size: 16000 words) which he hand-parsed using the state-transition formalism. On a test set of 100 sentences only 27 were given a fully correct analysis. This extremely modest result is partly due to the broad domain of the Brown corpus (cf. Francis & Kucera, 1982), of which only a relatively small fraction was analyzed by Tugwell. Furthermore, Tugwell only uses minimal corpus fragments, and as such his DOP model does not take maximally advantage of the statistical dependencies that may be found in his corpus. Tugwell's results are modest and preliminary, but his model is the first realization of the DOP approach for an essentially different theory of linguistic representation.

**A DOP model for LFG representations: Kaplan (1996)**

The tree representations of DOP1 ignore deep or functional syntactic structures that do not coincide with surface structures; moreover they do not even use syntactic features such as case, number or gender. Therefore, in Kaplan (1996) and Bod et al. (1996b), a DOP model is developed for the more articulated representations provided by Lexical Functional Grammar (LFG). On this theory of representation, the analysis of every utterance consists of a constituent-structure (a phrase structure tree), a functional-structure (an attribute-value matrix), and a correspondence function that maps between them (Kaplan & Bresnan 1982; Kaplan, 1995). Note that a constituent-structure in LFG corresponds to an utterance-analysis in DOP1. New are thus the functional-structure and the correspondence function.

As in the other versions of DOP, an analysis of a new utterance is generated by combining fragments from a corpus of utterance-analyses. A fragment of an LFG utterance-analysis is a triple $<c, \phi, f>$, where $c$ is a connected subtree of a constituent-structure, $f$ is a connected sub-functional-structure, and $\phi$ the correspondence function between them. Kaplan (1996) gives theory-based restrictions for LFG fragments: if a fragment includes a node $n$, it must include all of $n$'s sisters; if a fragment includes a node $n$ corresponding to sub-functional-structure $f$, then all other nodes under $n$ corresponding to $f$ must be included.

The combination-operator for LFG fragments uses DOP1's leftmost-substitution for combining sub-$c$-structures, extended with unification for the corresponding sub-$f$-structures. A derivation for an utterance can be generated by combining fragments such that the combination-operator applied from left to right results in a "valid" representation. The theory of LFG defines the notion of "valid": no non-branching dominance chains, a complete and coherent $f$-structure.

As to the probability model, the DOP1 method of normalizing fragment probabilities (by the number of fragments that can be combined with a previously constructed fragment), does not properly apply to LFG's nonmonotonic constraints on valid fragment combinations. For example, the completeness condition of LFG representations is something that can be evaluated only for a derivation as a whole, not at each stage of a derivation. Therefore, in Kaplan (1996) and Bod et al. (1996b), it is proposed to normalize the probabilities after the fact instead of beforehand. The probability of a derivation is initially computed as the product of the unnormalized fragment probabilities, which is then normalized by all valid fragment-sequences generating the same utterance. The probability of a representation is, like in DOP1, the sum of all derivations generating this representation.

English, French, German and Arabic corpora have been annotated with LFG representations, but no system for parsing with LFG fragments has been implemented yet.

## 7. Conclusion

We have seen that there is now a large and rapidly growing class of computational models that process new language input not by invoking the rules of a competence grammar, but by directly constructing analogies with concrete past language experiences. Independent experiments with different versions of this approach indicate that it disambiguates more accurately than other probabilistic methods.

Many of these experimental results have been achieved, however, with impoverished annotation formalisms (labelled phrase-structure trees), with very small corpora (ATIS), or with a sub-optimal version of the data-oriented approach (subtrees of depth one). Much work is still needed to be able to annotate large corpora with rich syntactic/semantic structures, and to develop parsing/disambiguation algorithms that can use such corpora in an efficient way.

To validate the psycholinguistic relevance of these models, they should be refined and extended in several ways. For instance, the parsing/disambiguation algorithms should be made to operate incrementally; recency and discourse structure should be brought into the picture; and generation algorithms should be developed.

The most challenging question about the data-oriented perspective on human language processing concerns the issue of language acquisition. If we want to avoid stipulating an "innate grammar", we must show how a corpus with syntactic structures may gradually come into being, starting with an initial state in which a child only has a corpus with non-linguistic experiences. We must show, therefore, how syntactic structures get bootstrapped by a process which (1) projects semantic/pragmatic structures onto the word sequences that accompany them, and (2) invents syntactic categories by noting semantic/pragmatic and distributional correspondences between syntactic constituents. Describing this process in any degree of detail will obviously be very difficult. Nevertheless, it constitutes a more promising research agenda than trying to find a connection between the observable child language acquisition data and the process of setting parameter values in an innate super-grammar.